\documentclass[pdflatex,sn-mathphys-num]{sn-jnl}
\usepackage{graphicx}%
\usepackage{multirow}%
\usepackage{amsmath,amssymb,amsfonts}%
\usepackage{amsthm}%
\usepackage{mathrsfs}%
\usepackage[title]{appendix}%
\usepackage{xcolor}%
\usepackage{textcomp}%
\usepackage{manyfoot}%
\usepackage{booktabs}%
\usepackage{algorithm}%
\usepackage{algorithmicx}%
\usepackage{algpseudocode}%
\usepackage{listings}%
\usepackage{soul}
\usepackage{hyperref}
\usepackage{natbib}

\definecolor{codegreen}{rgb}{0,0.6,0}
\definecolor{codegray}{rgb}{0.5,0.5,0.5}
\definecolor{codepurple}{rgb}{0.58,0,0.82}
\definecolor{backcolour}{rgb}{0.95,0.95,0.92}

\lstdefinestyle{mystyle}{
	backgroundcolor=\color{backcolour},   
	commentstyle=\color{codegreen},
	keywordstyle=\color{codegreen},
	numberstyle=\tiny\color{codegray},
	stringstyle=\color{red},
	basicstyle=\ttfamily\footnotesize,
	breakatwhitespace=false,         
	breaklines=true,                 
	captionpos=b,                    
	keepspaces=true,                 
	numbers=left,                    
	numbersep=5pt,                  
	showspaces=false,                
	showstringspaces=false,
	showtabs=false,                  
	tabsize=2
}

\usepackage{xcolor}
\colorlet{linkequation}{blue}

\newcommand*{\refeq}[1]{%
	\begingroup
	\hypersetup{
		linkcolor=linkequation,
		linkbordercolor=linkequation,
	}%
	(\ref{#1})%
	\endgroup
}

\hypersetup{
	colorlinks=true,
	linkcolor=black,
	urlcolor= blue,
	citecolor = codepurple}

\begin{document}

\title[Methods for PINNs]{Efficient PINNs via Multi-Head Unimodular Regularization of the Solutions Space}

\author*[1,2]{\fnm{Pedro} \sur{Tarancón-Álvarez}}\email{pedro.tarancon@fqa.ub.edu}

\author*[1,2]{\fnm{Pablo} \sur{Tejerina-Pérez}}\email{pablo.tejerina@icc.ub.edu}

\author[2,3]{\fnm{Raul} \sur{Jimenez}}\email{raul.jimenez@icc.ub.edu}

\author[4]{\fnm{Pavlos} \sur{Protopapas}}\email{pavlos@seas.harvard.edu}

\affil[1]{\orgdiv{Departament de F\'\i sica Qu\'antica i Astrof\'\i sica}, \orgname{Universitat de Barcelona}, \orgaddress{\street{Mart\'\i\  i Franqu\`es 1}, \city{Barcelona}, \postcode{ES-08028}, \state{Barcelona}, \country{Spain}}}

\affil[2]{\orgdiv{Institut de Ci\`encies del Cosmos (ICC)}, \orgname{Universitat de Barcelona}, \orgaddress{\street{Mart\'\i\  i Franqu\`es 1}, \city{Barcelona}, \postcode{ES-08028}, \state{Barcelona}, \country{Spain}}}

\affil[3]{\orgdiv{Instituci\'o Catalana de Recerca i Estudis Avan\c cats}, \orgname{ICREA}, \orgaddress{\street{Passeig Llu\'\i s Companys 23}, \city{Barcelona}, \postcode{ES-08010}, \state{Barcelona}, \country{Spain}}}

\affil[4]{\orgdiv{Institute for Applied Computational Science}, \orgname{Harvard University}, \orgaddress{\street{150 Western Ave.}, \city{Allston}, \postcode{MA 02134}, \state{MA}, \country{USA}}}

\abstract{Non-linear differential equations are a fundamental tool to describe different phenomena in nature. However, we still lack a well-established method to tackle stiff differential equations. Here we present a machine learning framework to facilitate the solution of nonlinear multiscale differential equations and, especially, inverse problems using Physics-Informed Neural Networks (PINNs). This framework is based on what is called \textit{multi-head} (MH) training, which involves training the network to learn a general space of all solutions for a given set of equations with certain variability, rather than learning a specific solution of the system. This setup is used with a second novel technique that we call Unimodular Regularization (UR) of the latent space of solutions. We show that the multi-head approach, combined with Unimodular Regularization, significantly improves the efficiency of PINNs by facilitating the transfer learning process thereby enabling the finding of solutions for nonlinear, coupled, and multiscale differential equations.}


\maketitle

\section{Introduction} \label{sec: intro}

Non-linear differential equations (DEs) describe complex systems in nature, such as fluid dynamics, climate modeling, and  general relativity, to mention a few. While linear equations are widely applicable, non-linear equations often arise in systems where interactions or feedback loops create more intricate and unpredictable behaviors. Yet, in most cases, analytic solutions are either impossible or highly complex, necessitating the use of numerical methods to provide solutions \cite{pinney1955}. With the advent of deep neural networks, particularly Physics-Informed Neural Networks (PINNs), it has become increasingly feasible to solve complex DEs more efficiently.
Extensive literature now exists on how to use PINNs to solve a wide variety of nonlinear, multiscale DEs including those in complex, real-world systems where traditional numerical methods may struggle. An added layer of complexity arises when the goal is to solve inverse problems, where the objective is to infer unknown parameters or functions within a system based on available data, such as initial or boundary conditions \cite{mpe_special_issue}.

Using PINNs to solve DEs offers several advantages over traditional methods. For instance, a PINN can be trained on a specific range of boundary conditions (BCs) or initial conditions (ICs). The specific solution(s) to a DE can only be determined if the corresponding ICs or BCs are specified as part of the problem. The existence and uniqueness theorem \cite{coddington1955theory}  ensures that there is a unique solution for a set of DEs once specific ICs or BCs are chosen. In traditional numerical methods, numerical integration is performed for each set of ICs-BCs. If multiple IC-BVs are given, the numerical integration must be run separately for each one. However, for PINNs, we can train the neural network (NN) on a specific set of DEs in a given range of IC-BVs referred to as a \textit{bundle}, which must be sampled \cite{flamant2020solving}. The training process may take more computational resources. However, once trained, the PINN can be evaluated almost instantaneously \cite{Chen2020, lu2021deepxde}  for any choice of IC-BV within the bundle, even if it was not trained specifically on it, and will still yield a solution. This may be extremely useful in cases where a very large number of solutions is required, each corresponding to different ICs-BCs or parameters \cite{karniadakis2021physics}.

Previous work has been done on using NNs for learning the operator mapping free functions $u(x)$, BCs or ICs given as inputs that appear explicitly in the DEs, to the corresponding solutions $s(x)$. The operators learned in DeepONets \cite{DeepONets_Lu_2021} and Fourier Neural Operators (FNO) \cite{FNO} can be used to predict new solutions $\tilde{s}(x)$ for stiffer input functions $\tilde{u}(x)$ that model has not trained on. In this sense, the end goal is similar to the one in this work: learning a higher-dimensional ``response'' of the NN (in our case, the latent space) that allows to compute solutions to different DEs by varying $u(x)$, BCs/ICs or parameters. However, the key difference stems in the fact that these other methods need to train on \textit{known solutions} $s(x)$ for each of the (random-)sampled $u(x)$ functions, computed from traditional numerical methods. In our approach, we directly solve the DEs, so no previous knowledge from any solutions is needed. In this way, the method discussed in the paper is a  different and complementary approach to DeepONets or FNOs.

Stiff DEs, e.g. DEs where there is a large separation of scales between different features of the solution, can be challenging to solve \cite{hairer1996solving, Seiler2025}. Stiffness can manifest in various forms, including eigenvalue-based stiffness (where the system's Jacobian has widely varying eigenvalues), multi-scale stiffness (involving both fast and slow time scales such as the flame equation (strictly speaking, the flame equation exhibits multiple types of stiffness) ), damping-induced stiffness (common in physical systems with strong damping such as the Van der Pol oscillator), nonlinear stiffness, interval-dependent stiffness, and transient stiffness \cite{lambert1991numerical}. In the cases we present, traditional numerical methods are capable of solving the equations. However, in the context of solving DE systems with PINNs, we present differential geometry techniques that significantly improve, or even unlock, our ability to solve them \cite{raissi2019physics}. Moreover, these techniques are particularly relevant when there is no established numerical setup to solve the problem. This is the case of inverse problems. PINNs are a well-suited tool for solving such problems \cite{lu2021deepxde}.

There are previous works where concepts of differential geometry have been used in the context of NNs. In \cite{8746812} the authors explore and characterize the behavior of NNs during training through the theory of
Riemannian geometry and curvature - they relate the output of the NN with the input transformations through the curvature equations. An analysis and comparison between the convergence properties of the metric of NNs (used for classification) and the metric under the Ricci-DeTurck flow is done in \cite{2021arXiv211108410C}. In \cite{Glass2020RicciNetsCP, inproceedings}, the authors remove edges of low importance (prune) the computational graph using the definition of Ricci curvature. More concretely, some recent works have proven that differential geometry techniques can increase the performance and accuracy of PINNs \cite{Arora2023,SahliCostabal2024,CardosoBihlo2025}.

Here, we present two efficient algorithms to use in PINNs that are aimed at significantly improving the solution of inverse problems for non-linear, coupled and multi-scale DEs. The first technique is called Multi-Head (MH) training, where we train the network to learn the general space of all solutions (or latent space) to different variations of the same DEs, from which we can perform transfer learning to different regimes \cite{MHPINNs, desai2022oneshot, zhou2024data, berardi2024inverse, 2023arXiv231114931L, DBLP:journals/corr/abs-2211-00214}. The second one is a technique that we call Unimodular regularization (UR) of the latent space of solutions. It is based on the computation of the metric tensor, a concept from differential geometry, induced on the latent-space-hypersurface. This second technique facilitates (or even unlocks) the transfer learning process to extreme regimes by ensuring a more controlled response of the latent space to changes in the DEs \cite{baldan2023physics}. The purpose of this article is to introduce these techniques and show their efficiency with several examples. The algorithms that we present here are of general use in solving DEs and inverse problems. The article is structured as follows: in the ``Methods'' section we introduce the methodology used; we describe the NN structure in ``Single head'' and ``Multi-head'' subsections, the general training procedure in ``Training and Transferring'' subsection, and UR in last subsection. The application of these techniques to different ordinary differential equation (ODE) systems and results are presented in the ``Results'' section, showing consistent improvement in the obtained solutions. We solve the Flame Equation, the Van der Pol Oscillator, and Einstein Field Equations in 5-dimensional Anti de Sitter spacetime in its corresponding subsections respectively. The conclusions are summarized in the ``Conclusions'' section.


\section{Methods} \label{sec: methodology}

\subsection{Context}
Physics-Informed Neural Networks (PINNs), first introduced in the works of Dissanayake and Phan-Thien \cite{dissanayake_neural-network-based_1994} and Lagaris et al. \cite{lagaris_neural-network_2000}, have emerged as a powerful framework for solving partial and ordinary differential equations (PDEs/ODEs). Raissi et al. \cite{raissi2019physics} popularized this approach by demonstrating its effectiveness on a variety of challenging physical problems, while other researchers have made significant advancements in applying neural networks to PDEs and ODEs, including notable examples such as \cite{mattheakis2021hamiltonian,sirignano2018dgm,zhu2019physics}. In this paradigm, one typically trains a distinct neural network for each unknown function in the governing equations. These neural networks are typically implemented as multilayer perceptrons (MLPs),  a class of models composed of stacked fully connected layers interleaved with nonlinear activation functions. The training process involves minimizing a loss function that encodes the squared residuals of these differential equations, thereby embedding the physics directly into the learning objective. For different approaches to embed the physics using machine learning techniques we refer the reader to the following references \cite{Choudhary2019,Choudhary2020,Greydanus2019}.

While traditional numerical methods often outperform PINNs in terms of computational efficiency and accuracy, PINNs offer unique advantages. 
One key benefit is their ability to learn the parametric dependence of solutions with respect to varying equation forms or initial/boundary conditions (IC/BC). In contrast, traditional methods typically require rerunning the solver for each set of different IC/BC. With PINNs, however, it is possible to train the network to simultaneously accommodate multiple conditions, enabling rapid evaluation of solutions for various IC/BC values or even different, parametrically related equations. By training on diverse initial and boundary condition samples, our approach can recover a free form function within the region covered by the observed data.  We caution, however, that accuracy is generally limited to the span of the conditions seen during training and may not extend reliably to entirely unobserved regions of the input space. However, in the third example of this work (subsection ``Einstein Field Equations'' in the ``Results'' section) the available initial and boundary condition samples sufficiently cover the region of interest, so the functional recovery remains accurate under this assumption.

Another notable advantage of using PINNs is their ability to leverage transfer learning. Transfer learning allows the model to generalize beyond the range of its training data, enabling the prediction of solutions in unseen regions or conditions. This technique has proven crucial for obtaining solutions in the stiffer regimes of certain differential equations \cite{DBLP:journals/corr/abs-2211-00214}. Throughout this paper, we will leverage this concept within the so-called MH framework.

The MH approach to solve DEs with PINNs aims to train a model that learns a general solution space for a given system, accommodating variations such as different ICs, BCs, coefficients/parameters, or forcing functions $f$ in the inhomogeneous part of the DE problem. Rather than learning a single specific solution, the model learns how the solution space behaves under all these different factors, allowing for generalization across a range of possible scenarios.

To formalize this approach, we represent the DE system as follows:
\begin{equation}
    \mathcal{D}\psi(x^\mu) - f(x^\mu) = 0\qquad ; \qquad\mu=1,2,3\dots
\end{equation}
where each $x^\mu$ represents a vectorial quantity of dimension equals to the batch size. $\mathcal{D}$ is a generic differential operator that encodes the form of the DE system, $x^1$ is the independent variable, $x^{2,3,...}$ a list of parameters describing the solutions for different ICs, BCs or parameters of the equations, and $\psi(x^\mu)$ is the solution of the system. $f(x^\mu)$ is a general function (often called the inhomogeneous part) that depends on the independent variable and could admit also a parameter.

From this point forward, we will refer to any scenario with the same DE form (same $\mathcal{D}$) and some choice of ICs, BCs, parameters ($x^{2,3,\dots}$) and/or forcing functions $f$ as belonging to the same \textit{family} of DEs. Any variation of $x^{2,3,\dots}$ and/or $f$ will be a variation within the family of the DEs, and a specific choice of them will be an \textit{element} of the family of DEs.

The multi-head (MH) approach is based on learning a mapping to a higher-dimensional latent space that goes beyond the specific solutions of the differential equations. 
In this higher-dimensional space, all potential solutions are encoded for variations within the family of DEs.  In the context of PINNs, this latent space is represented by a set of $d$ functions, $H_i(x^\mu)$, which are generated by the \textit{body} neural network (NN), whose output dimension is $d$. 

The latent space of solutions can then be “projected down” to a specific solution for a given element of the family of the DE system by non-linearly combining the $d$ components. This projection is performed by another NN, known as the \textit{head}. We will describe the multi-head setup in detail in the following subsections. For pedagogical clarity, we will first discuss a single-head model to explain its implementation within the PINNs framework. Subsequent subsections will show how this approach can be generalized to multiple \textit{heads} and how the latent space can be learned to extrapolate solutions to regions where the body has not been trained— a process known as transfer learning. Finally, we will introduce a technique called \textit{Unimodular Regularization}, which improves the performance of the MH approach when transferring solutions to stiff regimes of differential equations.

\subsection{Single-head} \label{subsec: single head}

\begin{figure}[H]
    \centering
    \includegraphics[trim=100 180 100 180, clip, width=\linewidth]{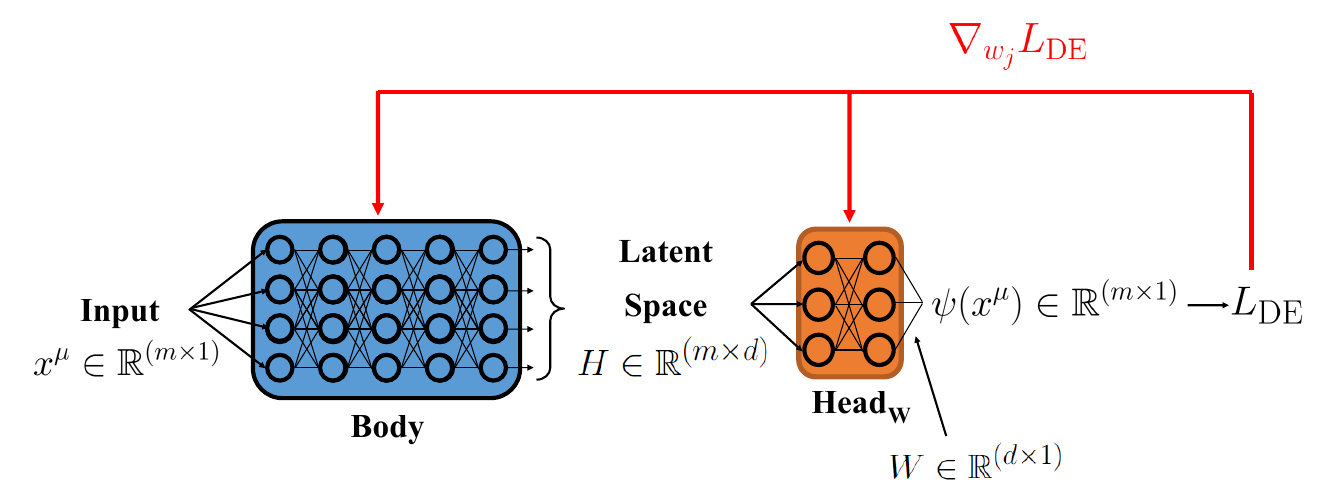}
    \caption{\textbf{Scheme of the single head approach Neural Network.} We can see how this simplified case is the standard case for a deep neural network with multiple hidden layers and one single output. The output solutions are calculated as $\psi^\text{NN} = \texttt{head}_W \left(H\right)$ with $\psi\in\mathbb{R}^{(m\times 1)}$, $H\in\mathbb{R}^{(m\times d)}$ and $W\in\mathbb{R}^{(d\times 1)}$, being $m$ the batch size (number of points sampled in the independent variable $x^1$ meshed with sampling of $x^{2,3,\dots}$), and $d$ the number of neurons in the last layer of the head. The loss $L_\text{DE}$ is of the form in Eq. \eqref{eq: loss function}, i.e. it contains the residuals of the DEs for which the NN is solving - this is the main characteristic of PINNs. The red arrow represents the updating of the NN-parameters $w_j$ (weighs and biases) based on minimization of the loss function through gradient descent (represented by $\nabla_{w_j} L_\text{DE}$).}
    \label{fig: single head}
\end{figure}

The \textit{single-head} approach involves using the latent space components $H_i(x^\mu)$ and passing them through a linear layer or another NN that produces the desired output dimension for the specific solution (in the case of a simple ODE, this would be a scalar output, i.e., the solution to the system). The structure responsible for processing $H_i(x^\mu)$ is called a \textit{head}, which projects the latent space into the specific solution:
\begin{equation}
     \psi(x^\mu) = \texttt{head}_W \left[H_i(x^\mu)\right]
\end{equation}
where $W$ represents the weights and biases of the \textit{head}, as shown in Fig. \ref{fig: single head}

In this simple case with a single head, we essentially have the standard setup of a NN with one output as the solution to an ODE problem.  
The parameters of the head and the body are initialized randomly and optimized by minimizing the loss using the Adam optimizer.
The loss function consists of the squared residuals of all DEs, averaged over the number of equations, sampled points along the independent variable, different ICs/BCs, parameters, and forcing functions. We can express the loss function for our PINN as follows:

\begin{equation}
    \label{eq: loss function}
    L_\text{DE}=\sum_{\mu=1}^n\sum_{\text{eqs}} \left[\mathcal{D}\psi^{\text{NN}}\left(x^\mu\right) - f(x^\mu)\right]^2
\end{equation}
where $\psi^{\text{NN}}$ is the solution predicted by the NN, and is guaranteed to satisfy the IC/BC by construction  (see ref. \cite{lagaris_neural-network_2000, DBLP:journals/corr/abs-2211-00214} for details).

The sum over $\mu$ represents the total over the different elements of the DE system, including  also all sampled points along the independent variable $x^0$.
Finally, the sum over the equations is the total of the residuals for each equation in our system of DEs. Note that the distinction between summing and computing the mean is irrelevant, as it merely scales the loss by a factor inversely proportional to the number of sampled points.

\subsection{Multi-head} \label{sec: multi-head}
We are now ready to understand how multiple heads enable the NN to learn the latent space of solutions across a whole family of DEs.

The goal is to find a specific solution $\psi_\alpha^\text{NN}(x^\mu)$ for each  problem by projecting $H$ into a reduced space, non-linearly combining its components through a head with parameters  $W_\alpha$, where the index $\alpha$ labels different variations of the DE family. Each head $W_\alpha$ correspond to a specific DE variation and produces a loss $L_{\text{DE},\alpha}$ of the form in Eq. \ref{eq: loss function}. We then define the total loss at each iteration as follows:
\begin{equation}
    L_\text{tot}=\sum_{\alpha=1}^{\#\text{heads}} L_{\text{DE},\alpha}
    \label{eq: MH total DE loss}
\end{equation}
and we perform gradient descent on the total loss. Thus, the MH PINN simultaneously solves all specified DE problems.

This approach uses one NN for each problem, with the \textit{body} shared across all NNs. Thus, the \textit{body} captures the global properties of the solutions, ideally representing the entire family of equations.

\begin{figure}[H]
    \centering
    \includegraphics[trim=20 10 15 10, clip, width=\linewidth]{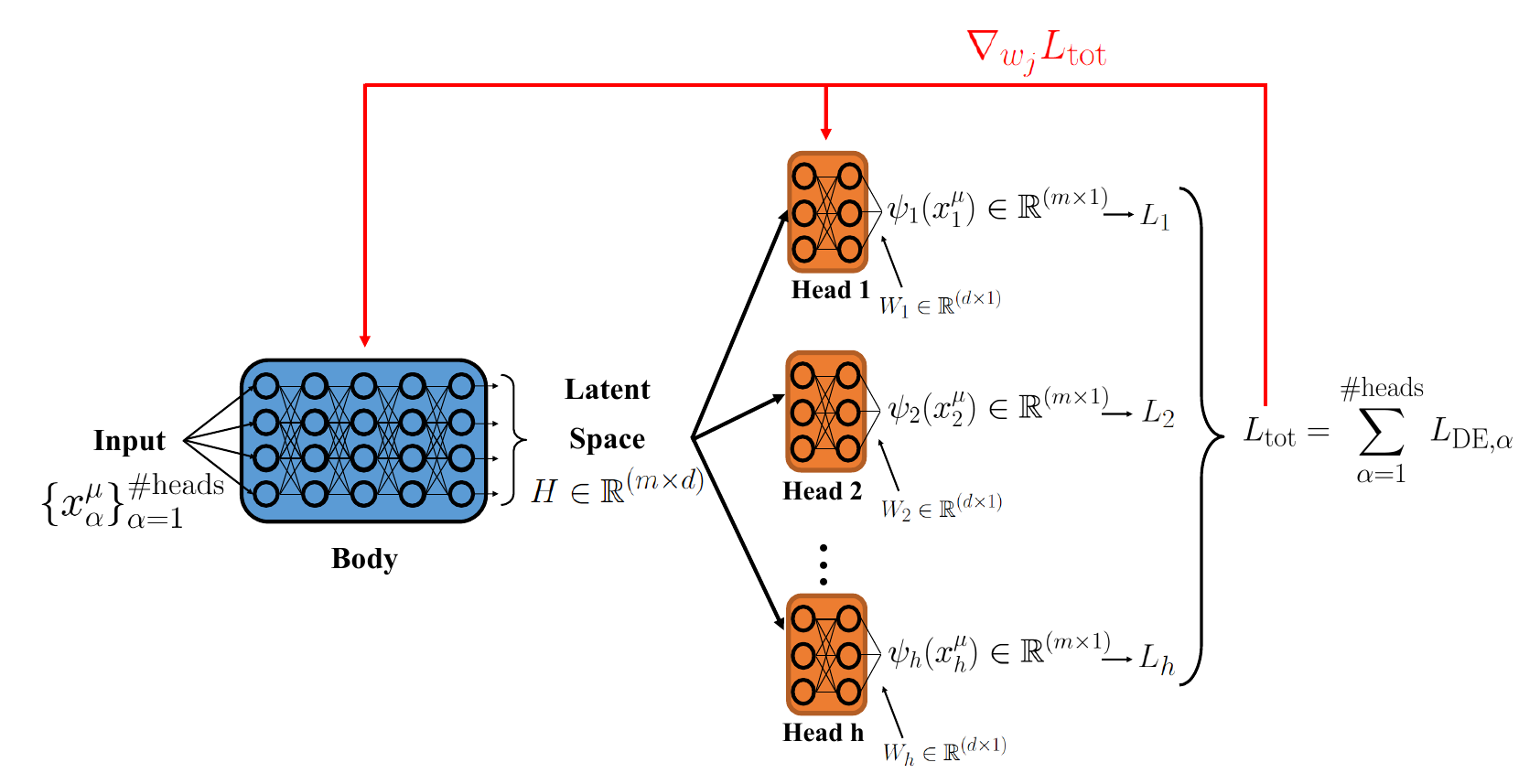}
    \caption{\textbf{Scheme of the multi-head approach Neural Network.} The index $\alpha$ in the inputs $x^\mu_\alpha$ denotes all the variations within the DE family - all of them are inputs to the \textit{body}, but each specific element of the family is later solved by its corresponding \textit{head}, giving solutions $\psi_\alpha(x^\mu_\alpha)$ as output ($\alpha=1,\dots,h$). The total loss $L_\text{tot}$ is as defined in \eqref{eq: MH total DE loss}. The backpropagation of the derivatives $\nabla_{w_j}L_\text{tot}$ indicated by the red arrow updates all the weights and biases $w_j$ (both of the \textit{body} and all \textit{heads}).}
    \label{fig: multi head}
\end{figure}

\subsection{Training and transferring} \label{sec: training and transfering}

The \textit{body} $H$ is trained using standard gradient descent techniques, with the loss function defined in Eqs. \refeq{eq: loss function} and  \refeq{eq: MH total DE loss}. As shown in the previous section and in Fig. \ref{fig: multi head}, each \textit{head} specializes in a specific variation within the family of DEs. If we continue training the \textit{body} and \textit{heads} with sufficiently diverse set DE elements, we expect the latent space $H$ to converge to a stable set of function values as the loss flattens.

Once the latent space $H$ stabilizes, we freeze the parameters of the \textit{body}, so they no longer change, even as the model continues training. We denote the frozen \textit{body} as $H^f$.

\begin{figure}[H]
    \centering
    \includegraphics[trim=100 180 100 180, clip, width=\linewidth]{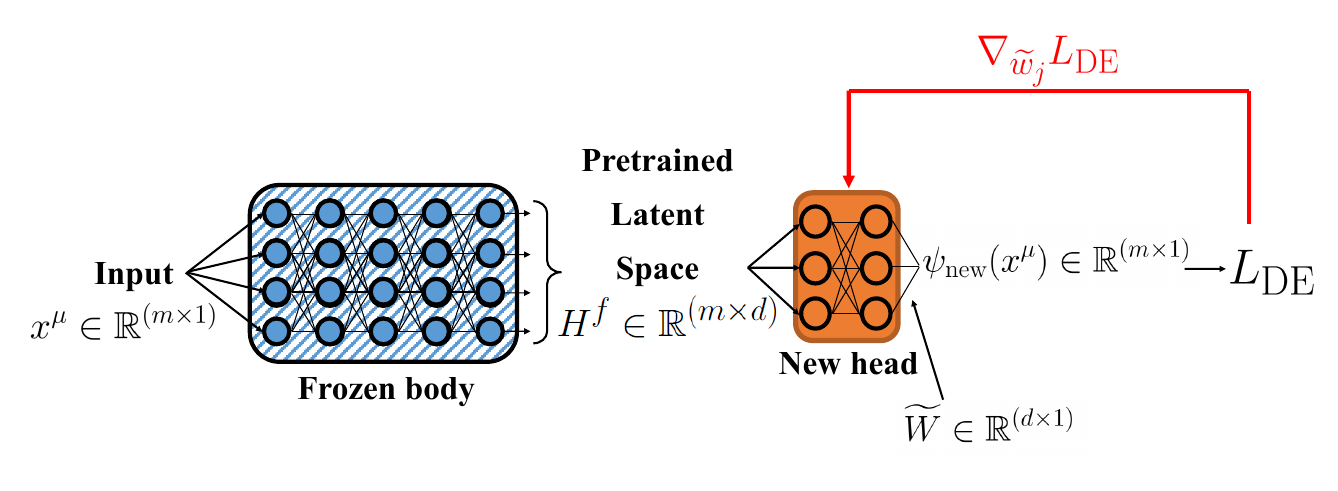}
    \caption{\textbf{Scheme of the transfer learning procedure from a pre-trained \textit{body}} \textbf{through the Multi-Head approach.} The blue, diagonal lines in the \textit{body} $H^f$indicate that it is frozen, i.e. its weights and biases do not change. Only the weights $\widetilde{W}$ of the new \textit{head} are learnt in order to find the solution to a new variation of the DE system, as indicated by the red arrow. As explained in the text, this method can be used both for getting solutions inside or outside the range in which the model has been trained on.}
    \label{fig: TL diagram}
\end{figure}

The pre-training of the latent space $H$ can be applied to solve any new variations of the DE family by simply attaching a new \textit{head} that takes $H^f$ as input. This process is referred as transfer learning (TL). The only part that requires training is the set of parameters $\widetilde{W}$ for the new \textit{head}, which is much smaller and computationally less expensive (see Fig. \ref{fig: TL diagram}):
\begin{equation}
    \psi^\text{NN}_{\text{new}} = \texttt{head}_{\widetilde{W}}\left(H^f\right).  
\end{equation}\\

The process of TL can be used in two different situations. The first one is to obtain solutions for new elements of the DE family given by choices of parameters/IC/BC/forcing functions whose values fall within the range of training of the \textit{body} $H^f$. The second is to find a specific solution to a variation within the DE family, even if that variation lies outside the values used in training. This is often the case when transferring to stiffer regimes, since it will be easier to train the system in the non-stiff regime. Next, we present a technique used during training and show that it improves results during the TL phase.

One might be concerned about the scalability of the multi-head approach to a large number of variations of the ODEs (or when generalizing to high-dimensional PDEs). A straightforward grid-based method needs \(\mathcal{O}(N^D)\) models for a \(D\)-dimensional space with \(N\) points per dimension, making it impractical for high \(D\). Conversely, our MH technique functions akin to a Monte Carlo sampler:
\begin{enumerate}
  \item \emph{Linear growth in heads.} We select \(N\) sample points in parameter space (e.g. stiffness parameter) and train one head per point, so the total number of heads grows as \(\mathcal{O}(N)\) rather than \(\mathcal{O}(N^D)\). Monte Carlo convergence theory ensures the approximation error decays as \(N^{-1/2}\), independently of \(D\).
\item \emph{Cost absorbed by the body.} For high-dimensional PDEs or bundle solutions, the main computational effort is sampling the expanded input space---this impacts the shared “body” during both forward and backward passes, not the number of heads itself.
  \item \emph{Comparison to VAEs.} Although a VAE’s encoder and decoder complexity scales with the latent dimension, VAEs still require dense input or parameter sampling for training. Our MH approach trades latent-space complexity for Monte Carlo–style coverage.
\end{enumerate}
In conclusion, MH requires a head for each sampled variation but mitigates the curse of dimensionality by utilizing Monte Carlo scaling (\(\mathcal{O}(N)\) heads for \(N\) samples), with exponential costs restricted to input sampling within the shared body, not through increased heads. Although we do not have a rigorous proof of Monte Carlo–style scaling, it is a well-justified assumption that has been observed empirically in many of our works.

\subsection{Unimodular regularization} \label{subsec: unimodular reg}

In this subsection, we introduce Unimodular Regularization (UR), a technique developed as part of the MH setup. This method involves imposing specific constraints on the values of the latent space and its derivatives, enabling the MH framework to generalize effectively across variations within the family of DEs. By regularizing the latent space in this manner, we enhance its adaptability and ensure robust transfer learning capabilities. The method’s theoretical foundations and technical implementation are detailed below.

As we have previously discussed, the \textit{body} NN in a MH setup is a critical component. Its primary function is to learn a generalized latent space of solutions, which serves as a functional basis for the problem. 
This latent space captures the underlying structure of the solution space of a family of DEs, and enables the model to represent a wide range of behaviors. The \textit{head} of the network then nonlinearly combines this latent space for a specific element of the DEs family to produce the corresponding solution. This separation between the generalized latent representation (\textit{body}) and specific parametrization (\textit{head}) ensures both flexibility and efficiency in solving the DEs.
We denote the latent space functions as $H_i$, where $i$ ranges from $1$ to $d$, with $d$ representing the dimension of the latent space. This latent space is determined by the inputs provided to the model, which we denote by $x^\mu$. Here, $\mu$ ranges from $1$ to $n$, where $n$ is the number of inputs of our model. 

Amongst these inputs, $x^1$ will always be the independent variable of the DE. The remaining variables,  $x^2, x^3, \dots, x^n$,  belong to parameter space and characterize different elements of the same family of DEs, and thus different solutions. These include variations due to different IC or BC, or specific parameter values or forcing functions within the DE itself. The exact nature of this parameter space depends on the problem being addressed.

During the training process, the \textit{body} of the MH will learn a latent space $H_i(x^\mu)$ that generalizes common features across different solutions to the DE within the same family. However, the choice of this set of functions is not unique, as the DEs are not \textit{exactly} solved during training (the DEs are solved by PINNs to the accuracy given by the square root of the loss function at a certain stage along the training), and the \textit{heads} simultaneously learn how to combine these functions to produce specific solutions non-linearly. This introduces a potential problem: some learned latent spaces may not generalize well to regions of the parameter space outside the training range. Specifically, if the latent space exhibits rapid variation with respect to the input parameters $x^\mu$, TL to new parameter ranges may become problematic. In such cases, the frozen latent space may take significantly different values for variations within the family of DEs that are close to each other. This often happens when the equations are stiffer in these ranges (this is typically the case in MH techniques, which are often used for TL in parameter ranges where the DEs are stiffer). The \textit{head}, in turn, would need to produce arbitrarily large responses to compensate for these rapid variations, making the solution unreliable.

To address this issue, we have developed a method called Unimodular Regularization (UR). This approach quantifies the values of the latent space and its derivatives by computing the associated metric tensor. By constraining the behavior of the latent space in this way, we ensure improved generalization and stability during TL, particularly in regions where the equations are more challenging. This is inspired by an alternative theory to General Relativity called Unimodular Gravity. In this theory the determinant of the metric tensor is picked to be $-1$ and the Gauge Group is reduced from diffeomorphisms (General Coordinate Transformations) to Transverse Diffeomorphisms (TDiff). These transformations are the ones that leave the determinant of the metric invariant. For more details see \cite{Alvarez:2023utn,Carballo-Rubio:2022ofy}.

We begin by using inputs $x^\mu$ as coordinates of a higher-dimensional manifold $\vec{\Omega}$ that also contains the latent space $H_i(x^\mu)$.
This is achieved by parametrizing the latent space as a hyper-surface of the inputs $x^\mu$. To formalize this, we group all components into the following quantity
\begin{equation}\label{eq: Omega}
    \vec{\Omega} \,=\, \left(x^\mu, H_i(x^\mu)\right)
\end{equation}
where $i=1,\dots, d$, and $\mu=1,\dots,n$. The total dimension of $\vec{\Omega}$ is given by $M = d + n$. Concretely, $H_i(x^\mu)$ is a $d$-dimensional hyper-surface parametrized by $n$ variables, embedded in an $M$-dimensional space. Using standard techniques from differential geometry, we can compute the induced metric tensor on the hyper-surface with the following expression:
\begin{equation}\label{eq: metric}
    g_{\mu\nu} \,=\, \frac{\partial \vec{\Omega}}{\partial x^\mu}\cdot \frac{\partial \vec{\Omega}}{\partial x^\nu}.
    \vspace{0.3cm}
\end{equation}
Here, $\cdot$ denotes the usual scalar product, defined as $\vec{x}\cdot \vec{y} = x_1 y_1 + x_2 y_2 + \hdots + x_ny_n$ where the dimension of each of the vectors is $n$. It is important to note that $g_{\mu\nu}$ is a matrix of dimension $n\times n$. Notably, as the metric tensor $g_{\mu\nu}$ is computed according to \refeq{eq: metric}, some entries of this matrix will be proportional to the derivatives of the latent space with respect to its inputs. These entries, therefore,  quantify the sensitivity of $H_i(x^\mu)$ to the inputs. 

To extract a scalar quantity from $g_{\mu\nu}$ that retains this sensitivity information we define the quantity $g$ as
\begin{equation}\label{eq: metric_det}
    g \,=\, \det g_{\mu\nu}
        \vspace{0.3cm}
\end{equation}
where $\det$ represents the determinant of the metric tensor. Note that, by definition, this quantity is always greater than or equal to  1.

Up to this point, we have introduced a method to use a scalar quantity, $g$, to quantify the sensitivity of the latent space to the model's inputs. Since our goal is to ensure this response is as smooth as possible, we now turn on how this can be imposed during  the model's training process. In the PINNs setup, the first step is to select a sampling of points for the independent variable and the parameters of the DE. The total number of points chosen is referred as the batch size. These inputs, denoted by $x^\mu$, constitute our training set. Once the sampling points are fixed, the response of the latent space $H_i(x^\mu)$ can be trivially evaluated at these points.

The implementation of equation \refeq{eq: metric} is relatively straightforward in the PINNs setup. In
order to compute the metric tensor it is necessary to compute the derivatives of the latent space with
respect to the inputs. These are computed analytically using automatic differentiation. This algorithm is
already implemented on most of machine learning pre-built packages. For the details of the implementation we refer the reader to our GitHub repository where all the codes can be found (see the ``Data and Code Availability'' section). This results in an $ n \times n$ matrix for each point in the batch, yielding a total size $=\text{batch size} \times n \times n $.
To simplify this structure, we use expression \refeq{eq: metric_det} to compute a scalar quantity for each point in the batch, reducing the dimensions to match the batch size. This scalar function quantifies the sensitivity, or rate of change, of the latent space with respect to the chosen input points.

To ensure that this rate of change remains smooth, we introduce an additional term to the loss function, denoted as  $L_\text{UR}$ , which takes the following form:

\begin{equation}\label{eq: add loss}
    L_\text{UR} \,=\, \lambda\sum_{\text{batch}}\left[\sqrt{g(x^\mu)} - \vec{\mathcal{I}}\right]^2.
\end{equation}
Here, $\vec{\mathcal{I}}$ is a vector of ones with a length equal to the batch size, and $\lambda$ is a free parameter that quantifies the relative weight of this term with respect to the differential equation loss. It is crucial to select $\lambda$ appropriately; if it is too large, the latent space response will become nearly constant, preventing the \textit{head} from projecting it to the desired solution. In our experiments, $\lambda$ was chosen to lie between $10^{-7}$ and $5\cdot 10^{-5}$, depending on the specific equations under consideration.

Although this regularization seems similar to Jacobian regularization (JR) technique \cite{jacobianreg}, it is fundamentally different in its geometrical interpretation; the determinant of the metric $g$ has a clear meaning in terms of differential geometry as the volume of a differential of the hyper-volume in which $H_i(x^\mu)$ lives. The pseudo-Jacobian present in JR lacks this interpretation. Also, we find that JR as a regularizing technique is implicit when performing UR. A detailed explanation can be found in the ``Supplementary Note 1''.

A qualitative explanation of the effect of adding the UR loss into the loss landscape is the following. We know that there must exist some combination of parameters (some state of the NNs) such that the NN output functions are a solution of the DEs up to some precession. This approximate solution corresponds to some local minimum in the high-dimensional space of parameters in which the loss landscape lives. However the ``throat'' that will bring the NN's state into this minimum is very narrow. We hypothesize that adding the UR loss \refeq{eq: add loss} has the effect of widening this throat. Then, it will be more likely for the NN to fall into the local minimum during the TL phase. This hypothesis is compatible with what we have observed in our tests, and with the losses shown in the figures of the following subsections.

The choice of using $\sqrt{g}$ is motivated by General Relativity. However, we expect to obtain similar results using $g$ instead of $\sqrt{g}$. Also note that because of the definition of the metric tensor, its determinant is always positive. Moreover, using this object rather than any other differential geometry quantities presents various advantages, among which we underline the following
    \begin{itemize}
         \item It is the simplest quantity one can compute. The determinant of the metric $g$ will only contain up to first derivatives of the latent space with respect to the inputs. Any other geometrical scalar of the manifold (for example, Ricci scalar) will imply computing second order derivatives of the latent space. During the tests, we found out that computing this quantity once can take up to $30$ minutes of computing time (this quantity needs to be computed recursively, which makes it unfeasible).
        \item The geometrical interpretability of this quantity is also clear: it measures the differential of volume of the latent space as a hyper-volume living in a higher-dimensional manifold parametrized by the inputs as coordinates. Consequently, the larger it is, the more latent space volume (or information) is captured within a coordinate differential.
    \end{itemize}

The interested reader can find a brief analysis of the relation of UR with the Lipschitz constant in the ``Supplementary Note 2'', as further motivation for the use of the metric determinant $\sqrt{g}$ in our regularizing technique.

Since computing $g$ is computationally intensive, we only considered this additional loss every $N$ epochs. For epochs that are not divisible by $N$, the term is set to zero. It is important for the stability of the model during training, specially when doing UR, to keep both the learning rate and the $\lambda$ parameter sufficiently small. The hyperparameter $N$ needs also to be chosen properly: if $N$ is too small, computational time increases, also the NN reduces its flexibility to fit solutions. If $N$ is too large, the regularization of the latent space is lost to all the training epochs without it. A proper $N$ will yield regularization that facilitates TL while keeping enough freedom of the NN to fit solutions. If the learning rate is very large the network will be ``pushed out'' of the local minima when the loss (9) is taken into account. We have found both the values for $\lambda$ and the learning rate by performing different tests of the algorithm. It is important to note that these values will change depending on the ODEs that one wants to solve. The specific values for each case are indicated in the ``Results'' section. Finally, if the model involves multiple bodies (as is often the case when the DE system includes more than one unknown function), the metric must be computed separately for each body. Consequently, for every unknown function in the system, a corresponding term of the form \refeq{eq: add loss} is added to the total loss. 

The overall scaling of computational time of the problem given UR is as follows: since we need to compute the derivatives in \eqref{eq: metric} through AD, training time scales as $b\times n \times d$ (being $b$ the batch size, $n$ number of inputs $x^\mu$, and $d$ dimension of the latent space). It scales linearly with the number of heads (each iteration implies a forward pass and a backward pass per head). It also scales linearly with the number of unknown functions in the DEs. It is complicated to give a quantitative statement of the scaling with the stiffness parameter, given that there is not a single definition of stiffness as described in the ``Introduction'' section. Qualitatively, as stiffness of the problem increases,the number of iterations also increases both for the MH body training, and the TL phase.

\section{Results} \label{sec: results}

\subsection{Application to three different system of ODEs}
In this section, we apply the proposed methods to three different ODEs of increasing complexity. First, we address the flame equation, a first-order, non-linear ODE that presents a challenge due to its rapid growth for specific initial conditions. Next, we tackle the van der Pol oscillator, a second-order, non-linear ODE that is reformulated as a system of two first-order ODEs. Finally, we will apply the method to an inverse problem: recovering a free function and solving a system of coupled, first-order, non-linear ODEs with different BC for the unknown functions. This problem, motivated by holography, involves solving the Einstein Field Equations (EFE) with a scalar field. 

We demonstrate that the proposed method successfully handles all three problems. Moreover, by imposing UR during the training of the body, the solutions obtained can be effectively transferred to more challenging regimes.

The structure of the following subsections will be consistent across the cases discussed. First, we train the \textit{body} of the MH setup using different \textit{heads}, each projecting the latent space toward the desired solutions for specific values of a given parameter or IC/BC (specific realizations of an element within the same family of ODEs). This training process will be performed twice: once with UR and once without it. Notably, the solutions obtained during this initial training phase appear identical in both cases.

Next, we will conduct TL by freezing the parameters of the \textit{body} and training a new \textit{head} for values of the parameter/IC/BC where the solutions are stiffer. 
In some cases, a proper solution through TL is only achievable when using a modified latent space geometry, which we refer to as the “regularized metric”. To highlight this, we will compare the performance of the standard MH setup (no UR) with the modified “UR MH” setup. This comparison shows that, under conditions of extreme stiffness, the MH wit UR setup facilitates TL much more effectively—or even exclusively—when the standard MH approach fails to converge to a solution. By incorporating UR, we further enhance the stability and generalization of the MH approach.

The results shown in this section were obtained by training the models on a single NVIDIA A30 graphic card. The total process of training the body and then learning a new head to transfer learn took few days in the most computationally demanding case (EFE).

In doing TL, we are reducing computational cost by leveraging the pretrained body (latent space representation). Since the parameters of the shared bodies are all frozen (i.e. do not change in the TL-phase), during TL, adapting to new regimes only requires training of the task-specific head. This approach reduces the computational cost of the solutions by producing each solution  as a projections of the pretrained latent space, rather than  retraining  both body and head  from scratch. 
Note that training the MH body can be costly, but it only has to be done once. Doing a quantitative analysis, we compare the time it takes to obtain a solution via training from scratch, or via projecting the latent space through a head (not considering the time taken in training the body). TL exhibits a speed increase of roughly $1.3$ times for the \textit{flame equation}, $1.5$ times for the \textit{VdP}, and $2.9$ times for the EFEs per solution. We refer the interested reader to \cite{DBLP:journals/corr/abs-2211-00214} for an analysis of the advantages of MH transfer leaning. We also note that because the body is frozen during TL, no backward pass is performed through it. As a result, we don’t need to store its activations, gradients, or optimizer states (first and second moments), yielding a roughly 6–8$\times$ reduction in memory footprint.

\subsection{Flame equation} \label{subsec: flame eq}
The flame equation describes the growth of a flame ball with radius $y$ as a function of the time $t$, starting from a fixed initial value $\delta$. The amount of consumed oxygen during combustion inside the ball scales with the volume $\sim y^3$ and the flux of incoming oxygen scales with the surface area $\sim y^2$. Different values of $\delta$ imply different realizations of the ODE as an element within the family of flame equations. This equation and the proper IC can be written in the following form
\begin{equation}\label{eq: flame equation}
    \frac{dy}{dt} \,=\ \rho(y^2-y^3);\quad y(0)\, =\, \delta: \quad 0\leq t \leq \frac{2}{\delta \rho}.
\end{equation}
where we have introduced an additional parameter $\rho$ that will artificially increase the stiffness of the problem. In our case we are choosing $\rho$ to be $300$. As we have mentioned, the solution to this equation will present a rapid growth on a time scale $2/\delta\rho$. This growth will be faster as we decrease $\delta$. Thus, the problem will be stiffer for smaller values of $\delta$.

\begin{figure}
    \centering
    \includegraphics[width = \textwidth]{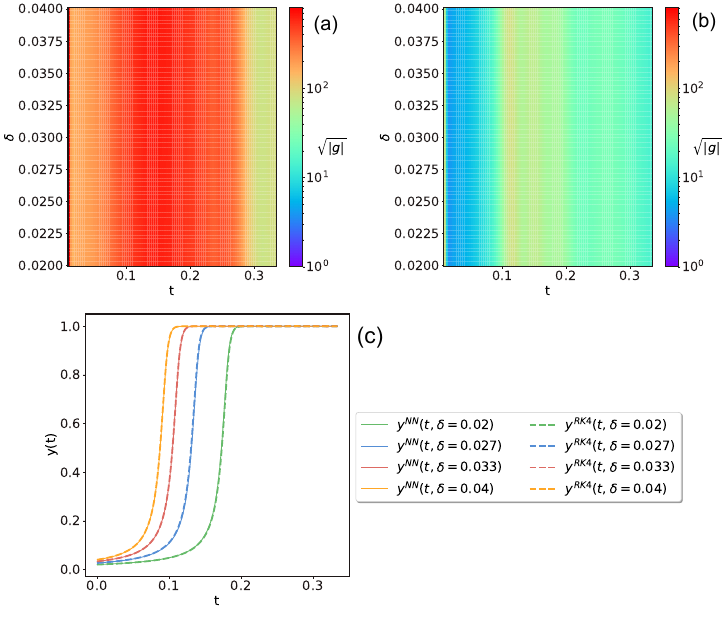}\vspace{-0.6cm}
    \caption{\textbf{Training of the Multi-Head \textit{body} for the flame equation}. $y^\text{NN}(t)$ is the ODE solution with independent variable $t$ given by the NN. $y^{RK54}(t)$ is the ODE numerical solution. $\delta$ is the initial condition on $y(t)$. $g$ is the value of the determinant of the metric. \textbf{(a)}: Metric determinant without UR. \textbf{(b)}: Metric determinant with UR. Note the difference in one to two orders of magnitude for the values of $g$ in each case. \textbf{(c)}: Solutions to the flame equation $y^\text{NN}(t)$ (solid lines) compared with the numerical-RK4 solutions (dahsed lines) for each of the \textit{heads}, for values of the parameter labeling different ICs, $y(0) = \delta \in [0.02, 0.04]$. As noted in the text, these solutions look identical for both cases with and without UR in the MH training process.}
    \label{fig:body_flame}
\end{figure}

The strategy for applying our method is as follows. We will first train two different \textit{bodies} using four \textit{heads} for $\delta \in [0.02, 0.04]$, for $1.5\cdot 10^{6}$ epochs. In one of them, we will impose UR as an additional loss. The results obtained for the solutions and the determinant of the metric are shown in figure \ref{fig:body_flame}. 

\begin{figure}
    \centering
    \includegraphics[width = \textwidth]{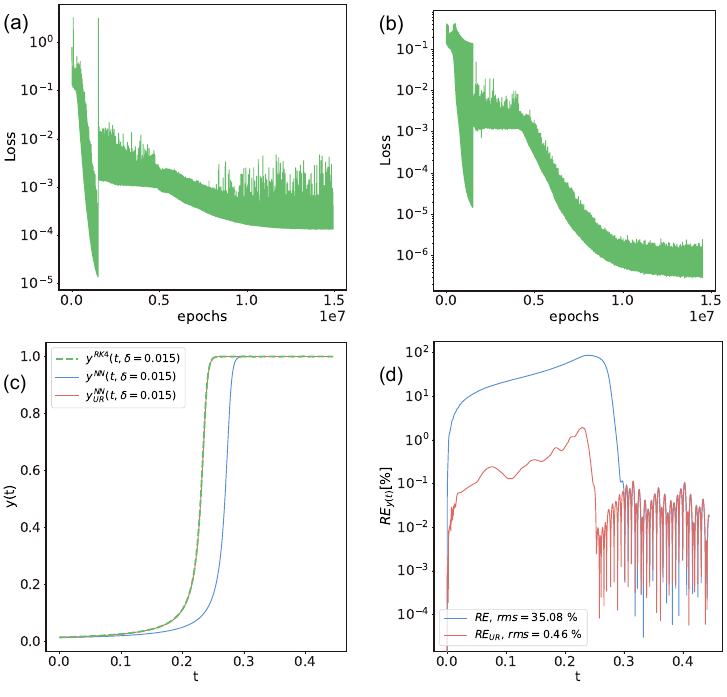}
    \caption{\textbf{Transfer learning for the flame equation}. $y(t)$ is the solution to the ODE, for numerical (RK4), and for NN with ($y_\text{UR}^\text{NN}$) and without UR ($y^\text{NN}$). Relative error (RE) between the numerical and the NN solution. The initial condition is $y(0)=\delta=0.015$. \textbf{(a)}: Loss functions for cases without UR. \textbf{(b)}: Loss functions for cases with UR. Transfer learning starts at epoch $1.5\cdot 10^6$. \textbf{(c)}: Solutions for $y(t)$ from a \textit{body} with UR (solid red line), without UR (solid blue), compared to the numerical solution (dashed green). \textbf{(d)}: RE (in \%) for cases with UR (red) and without UR (blue), as defined in Eq. \eqref{eq: relative error}. In the legend, we show the value of the root mean square (rms) for each case.}
    \label{fig:transfer_flame}
\end{figure}

Next, we will freeze the \textit{body} in both cases and perform TL by training a new head from scratch for a value of $\delta = 0.015$. The results of this TL are shown in figure \ref{fig:transfer_flame}, along with the relative error between the numerical and the NN solutions, computed as
\begin{equation}\label{eq: relative error}
    RE[\%] \,=\, \frac{\left|y^\text{RK4}(t) - y^\text{NN}(t)\right|}{1+ \left|y^\text{RK4}(t)\right|} \cdot 100
\end{equation}
where we have added the $+1$ factor in the denominator to avoid dividing by zero.

The model architecture and training details are summarized here. The \textit{body} is a Fully-Connected Neural Network (FCNN) with two input units, 64 output units, and three layers of 64 neurons each. Each \textit{head} is an FCNN with $64$ input units, one output, and two layers of 32 neurons each, using \texttt{Tanh} as the activation function.  For MH \textit{body} training, we used \texttt{adam} as the optimizer with an initial learning rate of  $1 \cdot 10^{-3}$ , reduced by 5\ every 15,000 epochs using a step scheduler. TL followed the same parameters for both bodies but used a higher learning rate ( $5 \cdot 10^{-3}$ ), reduced by 2.5\% every 2,500 epochs. 

For training, we sampled 100 points in the  $t$ -direction between the time bounds in \refeq{eq: flame equation}, adding small random noise at each epoch to prevent overfitting. When training the MH \textit{body}, we used  $\delta = 0.02$  (the smallest value in the range) to ensure the widest time intervals. For UR, we set  $\lambda = 5 \cdot 10^{-7} $ and imposed the additional loss term every 100 epochs.

\subsection{Van der Pol oscillator} \label{subsec: VdP eq}

\begin{figure}
    \centering
    \includegraphics[width = \textwidth]{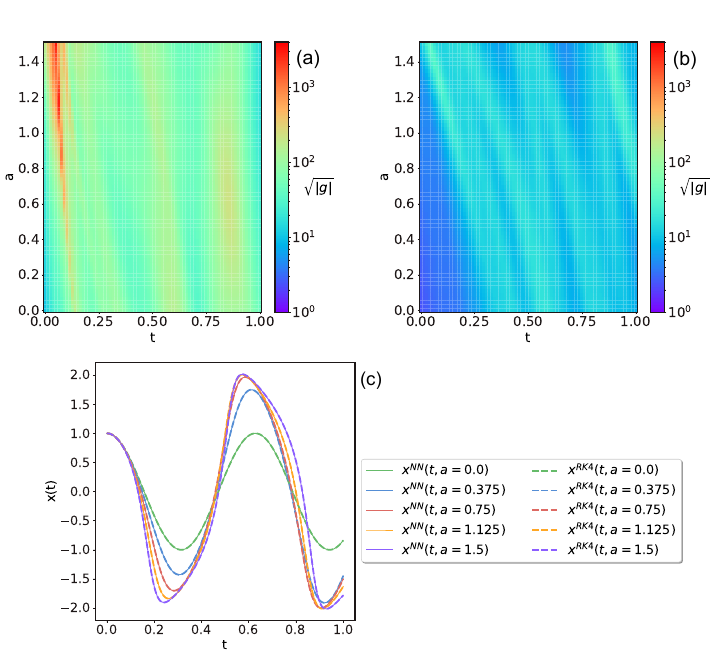}
    \caption{\textbf{Training of the multi-head \textit{body} for the van der Pol oscillator}. $x(t)$ is the solution to the ODE numerical (RK4) and given by the NN. $a$ is the stiffness parameter appearing in the ODE and $g$ is the determinant of the metric. \textbf{(a)}: metric determinant of the $y(t)$ \textit{body} without UR. \textbf{(b)}: metric determinant of the $y(t)$ \textit{body} with UR. \textbf{(c)}: solutions to the van der Pol oscillator $x^\text{NN}(t)$ (solid lines) compared with the numerical-RK4 solutions (dashed lines) for each of the \textit{heads}, for values of the parameter $a \in [0.0, 1.5]$. The solutions look identical for both cases with and without UR in the MH training process.}
    \label{fig:VdP_body}
\end{figure}

\begin{figure}
    \centering
    \includegraphics[width = \textwidth]{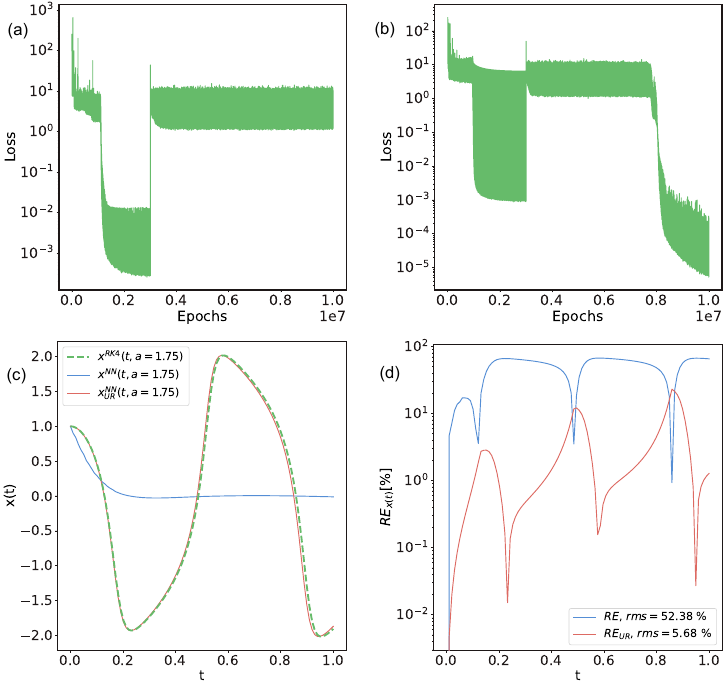}
    \caption{\textbf{Transfer learning of van der Pol oscillator}. $x(t)$ is the numerical (RK4) and NN solution to the ODE. The stiffness parameter appearing in the ODE, for this case of TL $a=1.75$. RE is the relative erorr between the numerical the NN solutions. \textbf{(a)}: Loss functions for the TL without UR. \textbf{(b)}: Loss functions for the TL with UR. TL starts at epoch $3\cdot 10^6$. \textbf{(c)}: NN-solution $x^\text{NN}(t)$ for cases without UR (solid blue), with UR (solid red) compared to the RK4-numerical solution (dashed green). \textbf{(d)}: Relative error (in \%) for cases with UR (red) and without UR (blue). In the legend, we show the value of the root mean square (rms) for each case.}
    \label{fig:transfer_VdP}
\end{figure}

The van der Pol oscillator represents a specific instance of a non-linear, damped harmonic oscillator. It can be mathematically characterized by the subsequent second-order ODE along with the IC:
\begin{equation}\label{eq: VdP}
    \frac{d^2 x}{dt^2} - a (1-x^2) \frac{dx}{dt} +  x\,= \,0; \qquad x(0) \,=\, 1; \qquad\left. \frac{dx}{dt}\right|_{t=0} \,=\, 0,
\end{equation}
where $x(t)$ can be understood as the oscillator's position over time. In this context, $a$ is regarded as a damping coefficient and modulates the problem's stiffness. The greater its value, the more challenging the equation will be to resolve. A key distinction between this equation and the preceding case is that it represents a second order ODE. Therefore, for the PINNs method to efficiently converge to the desired solution, it is advantageous to transform equation \refeq{eq: VdP} and the IC into a system of two first order ODEs by introducing an auxiliary unknown function $y(t)$.
\begin{equation}\label{eq: VdP system}
    \left\{
        \begin{aligned}
            \frac{dx}{dt} \,&=\, \rho y\\
            \frac{dy}{dt} \,&=\, \rho a(1-x^2)+ \rho x
        \end{aligned}
    \right.\qquad \text{IC: }
     \left\{
        \begin{aligned}
            x(0) \,&=\, 1\\
            y(0) \,&=\, 0.
        \end{aligned}
    \right.
\end{equation} 
As in the previous case, we have additionally introduced $\rho$ as a parameter to artificially increase the stiffness of the problem in the domain of interest. Throughout this subsection, we will fix it at $10$.

The primary distinction between this case and the previous one lies in our treatment of $x$ and $y$ as independent functions. Consequently, we introduce an additional NN to serve as the extra function. The MH strategy will largely remain unchanged, except for the necessity of training one more NN. Initially, as before, we train the main network using five distinct \textit{heads}, each responsible for values of $a$ within the interval $[0.0, 1.5]$. We conduct this training over $3 \times 10^6$ epochs for two separate main networks. In one of this set of networks, we apply UR as an extra loss term. The derived solutions, along with the determinant of the metrics for both networks, are displayed in Figure \ref{fig:VdP_body}. Subsequently, the main networks are fixed, and a new \textit{head} is trained over $7 \times 10^6$ epochs to facilitate TL to $a = 1.75$. The TL results for both networks, along with their relative error—which is calculated using equation \refeq{eq: relative error}—are illustrated in Figure \ref{fig:transfer_VdP}.

This paragraph outlines the key aspects of both the architecture and the training methodology. In constructing the MH \textit{body}, we employed two fully connected neural networks (FCNN), each having two input units and an output size of 128, with four layers consisting of 64 neurons each. The \texttt{Tanh} function was selected as the activation function for both networks. For the \textit{heads}, we selected an FCNN with 128 input units, a single output dimension, and two layers, each containing 64 neurons. The training of the body involved the \texttt{adam} optimizer, initialized with a learning rate of $10^{-3}$. A learning rate scheduler was applied to decrease it by $2.5\%$ every $15,000$ epochs. For TL, we also used the \texttt{adam} optimizer, starting with an initial learning rate of $10^{-3}$, and a scheduler decreasing it by $4\%$ every $70,000$ epochs. The training dataset consisted of $100$ evenly distributed points in $t$ ranging from 0 to 1. To prevent overfitting, random noise was added to these points at each epoch. A UR penalty, with $\lambda = 4 \cdot 10^{-5}$, was applied every $100$ epochs as an additional loss.

\subsection{Einstein Field Equations in AdS background} \label{subsec: EFE holo}

Before going into details of this case, it is important to note that the level of difficulty here is significantly higher than in the previous examples. The  ODEs  considered earlier served mainly  as toy models to illustrate both the MH setup and UR technique, and their solutions and behaviors are well understood.  However, the example presented in this section has only been solved for not very stiff regimes \cite{Bea:2024xgv}. 

Moreover, its resolution involves solving a system of seven ordinary highly non-linear ODEs for approximately $70$ BCs, and recovering a free-form function appearing within the equations that is uniquely determined by these BCs. Consequently, this stands out as one of the most challenging setups for testing the algorithms introduced so far. 

As previously indicated, our system comprises seven first-order, non-linear ODEs \refeq{subeq: E1}--\refeq{subeq: E7}, with six being linearly independent. At first glance, one can observe that the first three equations are simply redefinitions of the derivatives of the variables $\tilde{\Sigma},\,\tilde{A}$ y $\phi$. This arises because the Einstein Field Equations (EFE) are second-order differential equations, which have been rewritten as first-order equations. For a detailed discussion on the physical motivation behind these redefinitions, the reader is referred to \cite{Bea:2024xgv} and references therein.

\begin{subequations}
\label{1s_order_ODEs}
\begin{eqnarray}
&& \nu_{\Sigma}  - \tilde{\Sigma}'  = 0 \,, \label{subeq: E1} \\[2mm]
&& \nu_A  - \tilde{A}' = 0 \,, \label{subeq: E2} \\[2mm]
&& \nu_\phi  - \phi'  = 0 \,, \label{subeq: E3} \\[2mm]
&& \nu_{\Sigma}' +  \frac{2}{3} \tilde{\Sigma} \, 
 \nu_{\phi}^2  = 0 \,, \label{subeq: E4} \\[2mm]
&& u^2 \, \tilde{\Sigma} \, \nu_A^{\prime} + \frac{8}{3} \, V(\phi)\, \tilde{\Sigma}+ 
    \nu_A \, \left(  3u^2 \, \nu_{\Sigma} - 5u\, \tilde{\Sigma}  \right) + 
           \tilde{A} \left( 8\tilde{\Sigma} - 6u \,\nu_{\Sigma} \right) = 0 \,, \label{subeq: E5} \\[2mm] 
&& u^2 \,\tilde{\Sigma} \, \tilde{A}\, \nu_\phi' - \tilde{\Sigma}\,  \frac{d V}{d \phi} 
        + \nu_{\phi} \left( -3 u \, \tilde{A} \, \tilde{\Sigma}  + u^2 \,\tilde{\Sigma} \,\nu_A 
        + 3 u^2\, \nu_{\Sigma} \, \tilde{A} \right)  = 0 \,,  \label{subeq: E6} \\[2mm]     
&& \left( u \, \nu_{\Sigma}-\tilde{\Sigma} \right) 
\left( u^2 \, \tilde{\Sigma} \, \nu_A + 2 u^2  \, \tilde{A} \, \nu_{\Sigma} - 4 u \tilde{A} \tilde{\Sigma} \right)
- \frac{2}{3}  u \, \tilde{\Sigma}^2 
\left( u^2 \tilde{A} \, \nu_\phi^2 - 2 V(\phi) \right)  = 0\, \label{subeq: E7} . \,\,\,\,\,\,\,\,\,\,       
\end{eqnarray}
\end{subequations}

The independent variable is $u$. The six functions (dependent variables) are   $\Vec{\psi}(u)\equiv~\left( \tilde{\Sigma}(u),\tilde{A}(u),\phi(u),\nu_\Sigma(u),\nu_A(u),\nu_\phi(u) \right)$. Looking at Eqs. \refeq{subeq: E5}--\refeq{subeq: E7}, we can see a free function $V(\phi)$ and its derivative. Evidently, for different $V(\phi)$, one would get different equations, and thus different solutions $\Vec{\psi}(u)$.
    
The BCs are given by:

\begin{subequations}
    \label{BoundaryConditions}
    \begin{align}
    \tilde{\Sigma}|_{u=0} &=1 \,\,, \label{BoundaryCondition_a} \\[1mm] 
    \tilde{A}|_{u=0} &=1 \,\,,  \label{BoundaryCondition_b}\\[1mm]
    \phi |_{u=0} &=0 \,\,,  \label{BoundaryCondition_c}\\[1mm]
    \nu_\phi|_{u=0} &= 1 \,\,, \label{BoundaryCondition_d}\\[1mm]
    \tilde\Sigma_{u=1} &= \left( S/\pi \right)^{1/3}\,\,, \label{BoundaryCondition_e}\\[1mm]
    \tilde{A}|_{u=1} &=0 \,\,,  \label{BoundaryCondition_f}\\[1mm]
    \nu_A |_{u=1} &= - 4 \pi T \,\,. \label{BoundaryCondition_g}
    \end{align}
\end{subequations}

The BC for $\tilde{\Sigma}(u)$ and $\nu_A(u)$ at $u=1$ depends on a pair of values $\{T,S\}$ (temperature and entropy). It should be noted that there is an additional boundary condition present. Since we are not solving the equations exactly, giving additional constraints (when available) is helpful for the learning process.

The physical reasoning associated with this system primarily clarifies two aspects. Firstly, there will be a specific relation between temperature and entropy $S=S(T)$, meaning that for a choice of $T$, the value of $S$ is fixed. Secondly, there is an exact one to one correspondence between the free function $V(\phi)$ and the relation $S(T)$.\\

\noindent In this setting, we define the {\em direct problem} as follows:
\begin{enumerate}
\item \textbf{Select a specific $V(\phi)$: }
the function $V(\phi)$ must satisfy certain constraints on its value and derivatives near $\phi=0$ to ensure the resulting equations are well-defined. For a detailed discussion, see \cite{Bea:2024xgv, Bea:2018whf}. A chosen $V(\phi)$ specifies an entire family of ODEs, with different BCs giving rise to variations within that family.
\item \textbf{Choose $N$ distinct sets of boundary conditions,} $\{\text{BC}_i\}_{i=1}^N$.

\item \textbf{Numerically solve the ODE systems} defined by Eqs.~(\ref{subeq: E1})--(\ref{subeq: E7}) together with each of the $\text{BC}_i$, using standard methods (e.g., \texttt{NDSolve} in \emph{Mathematica}, see \cite{Bea:2018whf,Gubser:2008ny}). Denote the corresponding solutions by $\{\vec{\psi}_i(u)\}_{i=1}^N$.

\item \textbf{Evaluate each solution} $\tilde{\Sigma}_i(u)$ and $\nu_{A,i}(u)$ at $u=1$ to obtain $S_i$ and $T_i$ via Eqs.~(\ref{BoundaryCondition_e}) and (\ref{BoundaryCondition_g}), respectively. By pairing each $T_i$ with its corresponding $S_i$,  one obtains the $S(T)$ curve associated with the initial choice of  $V(\phi)$
\end{enumerate}

Note again that the direct problem can be handled with traditional numerical methods.

Let us now discuss the \textbf{inverse problem}: we take a sample of \(N\) points from a chosen \(S(T)\) relation, which specifies a bundle \(\{\text{BC}_i\}_{i=1}^N\). Consequently, the “full solution” \(\{\vec{\psi}_i(u)\}_{i=1}^N\) becomes a set of \(N\) different solutions to Eqs.~\refeq{subeq: E1}--\refeq{subeq: E7} that must be found. Even though we know there is a corresponding function \(V(\phi)\) related to the chosen \(S(T)\)/BCs, we do not actually know this function; it remains a free function determined by the solutions \(\phi_i(u)\in\vec{\psi}_i(u)\) for each \(i=1,\dots,N\).

The task is \textbf{finding the unique free function \(V(\phi)\)} such that the corresponding ODE family admits solutions \(\{\vec{\psi}_i(u)\}_{i=1}^N\) \textbf{for all \(\{T_i,S_i\}\) in the bundle}, while simultaneously determining those solutions. Although each \(\{T_i,S_i\}\) (and thus each set of boundary conditions) leads to a different solution \(\vec{\psi}_i(u)\), all solutions must share the same free function \(V(\phi)\).

Solving this inverse problem with traditional methods is not possible, because the equations must be fully specified before numerical integration can proceed---they cannot depend on \(V(\phi)\) when it is treated as a free function.

In \cite{Bea:2024xgv}, we and our collaborators developed an algorithm that uses PINNs to solve this problem. To check that the algorithm worked, we followed the work by \cite{Bea:2018whf}, where a theoretical potential $V^\text{th}(\phi)$ is chosen, and the direct problem is solved numerically to obtain $S^\text{th}(T)$. Then, we solved the inverse problem, giving this $S^\text{th}(T)$ as input to our PINNs setup as BCs. We find the solutions and recover the free function, $V^\text{PINN}(\phi)$. In this way, we are able to compare the original and recovered free functions $V^\text{th}(\phi)$ and $V^\text{PINN}(\phi)$, respectively. Also, by solving numerically once more the direct problem using now the recovered $V^\text{PINN}(\phi)$, we obtained the associated $S^\text{PINN}(T)$, which can again be compared with the original $S^\text{th}(T)$.

However, besides being a challenging problem due to its inverse nature, there are several other complications. The primary complication arises from the morphology of \( S(T) \), which can induce features in the associated \( V(\phi) \) at vastly different scales. This separation of scales significantly increases the stiffness of the problem. {\em In general, when a system exhibits features across multiple scales, numerical methods must accommodate the rapid variations alongside the slower ones, leading to stiffness}. In our case, we quantify the complexity of the morphologies of \( S(T) \), and thus indirectly the stiffness of the problem, using a single parameter \( \phi_M \). Specifically, we operate within the model selected in \cite{Bea:2018whf}, where the parameter \( \phi_M \) is integral to the choice of \( V(\phi; \phi_M) \) when solving the direct problem. Lower values of \( \phi_M \) correspond to richer morphologies of the associated \( S(T) \), introducing more pronounced scale separations.

Subsequently, we tackle the inverse problem using selections of \( S(T) \) with known morphological relationships to \( \phi_M \). While a general choice of \( S(T) \) might not be governed by a single parameter, within the same physical context, the correspondence between \( S(T) \) and \( V(\phi) \) remains intact. It is important to note that the link between scale separation and stiffness, while observed and utilized in our framework, is not derived from a formal mathematical proof but rather from empirical and heuristic understanding. Therefore, even though the stiffness of our EFE problem is influenced by the morphology of \( S(T) \), we refer to \( \phi_M \) as a ``stiffness parameter'' analogous to the parameter \( \delta \) in the flame equation \refeq{eq: flame equation} and the parameter \( a \) in the Van der Pol (VdP) equation \refeq{eq: VdP}. Lower values of \( \phi_M \) exacerbate the problem's stiffness, making it increasingly difficult to recover the free function. This difficulty persists until the original PINN setup either fails to recover \( V(\phi) \) within an acceptable error margin or, in cases of extreme stiffness, fails to recover it altogether, including its most fundamental features, such as the position \( \phi_H \) and scale \( V(\phi_H) \) of the critical points. For a more in-depth discussion on the stiffness of the problem, the features in \( V(\phi) \), and the morphologies of \( S(T) \), we refer the reader to Figures 1, 2, and 13 of \cite{Bea:2024xgv}.

In an attempt to invert this problem given data $S(T)$, one might be concerned that the sampled data does not cover the whole observable space, and thus the inverted function $V(\phi)$ cannot be fully recovered. However, this is not the case in this setup. The “observed data” \(S(T)\) are in fact synthetic, obtained by numerically solving the direct EFE problem for a potential \(V\) drawn from the same family \(V(\phi, \phi_M)\) at a fixed \(\phi_M\). Each sample in \(S(T)\) corresponds to one set of BCs, fully defining the solution and thus restricting the solution space to a single solution. We further truncate \(S(T)\) at very high temperatures---an approximation justified by the problem’s symmetry, since beyond the asymptotic regime (\(S\propto T^3\); see \cite{Bea:2024xgv}), no new information is gained.  Moreover, the sampled data for the independent variable $u$ will cover most of the observable space. It ranges from $0$ to $1$ which in the physical context correspond to the whole size of the spacetime in which EFEs are being solved. It is always sampled using a \textit{Chebyshev} distribution with noise. Thus, we expect that this range will be entirely covered when the model is trained for a sufficiently large number of iterations.

\begin{figure}[H]
    \centering
    \includegraphics[width = \textwidth]{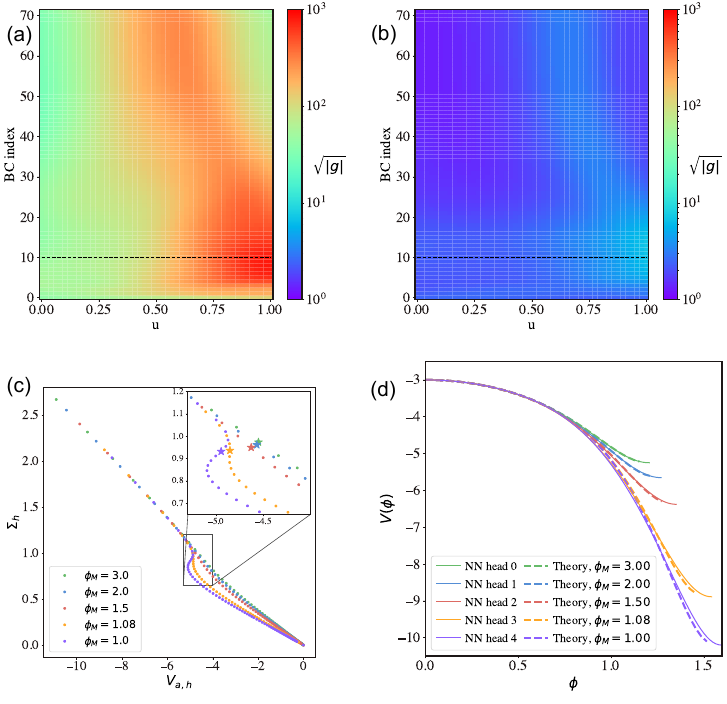}
    \caption{\textbf{Training of the multi-head body for the Einstein Field Equations.} g is the metric determinant. $\phi_M$ is the stiffness parameter. $\phi$ is one of the unknown solution functions, and $V(\phi)$ the unknown potential. $\Sigma_h$ and $V_{a,h}$ are different boundary conditions of the unknown functions. \textbf{(a)} Metric determinant for cases without UR. \textbf{(b)} Metric determinant for cases without UR. The head corresponds to the parameter $\phi_M=1$, for one of the solution functions, $\nu_\Sigma(u) \in\vec{\psi}(u)$.  
    In both panels, the black, dashed, horizontal line corresponds to the ODE problem given by the BC that is marked with a purple star in panel (c) (theory with $\phi_M=1$). \textbf{(c)}: Different input points given to the body that come from different choices of $S(T)_{\phi_M}$. Marked with a star, we show the BC point with index $i=10$ for all $\phi_M$. This point approximately tracks where the equations become stiffer within the corresponding $\phi_M$ family. \textbf{(d)}: Recovered NN potentials $V^\text{NN}(\phi;\phi_M)$ coming from the MH \textit{body} are shown in solid lines, while the theoretical $V^\text{th}(\phi;\phi_M)$ are shown in dashed lines.}
    \label{fig: body holo eqs}
\end{figure}

In this context, we introduce a \textbf{MH approach}: we supply our MH PINN framework with five increasingly stiff curves \(S(T)_{\phi_M}\), corresponding to
\(\phi_M = \{3.0,\,2.0,\,1.5,\,1.08,\,1.0\}\) 
(see Fig.~\ref{fig: body holo eqs}, bottom panels). Through its shared \textit{body}, the MH PINN learns the latent space of solutions and free functions associated with the various morphologies of \(S(T)_{\phi_M}\).
Subsequently, each \textit{head} projects down from this shared latent space to produce a specific
\(V^\text{PINN}(\phi; \phi_M)\),
as well as the corresponding ODE solutions $\Vec{\psi}_{i,\phi_M}$,
for the BCs specified by each 
\(\{T_i,S_i\}_{\phi_M}\).

\begin{figure}[H]
    \centering
    \includegraphics[width = \textwidth]{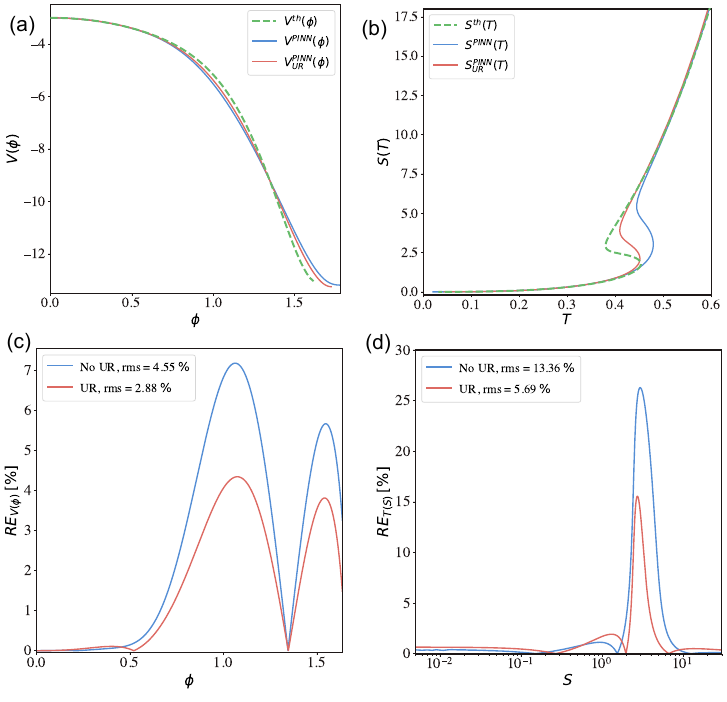}
    \caption{\textbf{Transfer learning of Einstein Field Equations for a non-pronounced first order phase transition}. $V(\phi)$ is the free form potential recovered by the NN (PINN) and the theoretical one (th). $S(T)$ is the thermodynamical curve (entropy as a function of temperature) recovered by the NN (PINN) and the theoretical one (th). RE refers to relative error between the NN model and the theoretical result. \textbf{(a)} Recovered functions $V^\text{PINN}(\phi)$ through TL without UR (solid blue), and with UR (solid red), compared with the theoretical function (dashed green), corresponding to $\phi_M=0.9$. \textbf{(b)} $S^{\text{PINN}}(T)$ relations associated to the recovered $V^\text{PINN}(\phi)$ without and with UR, compared to the theoretical one (which is also the initial input in the TL procedure). \textbf{(c)}: REs for the recovered $V^\text{PINN}(\phi)$ for the cases without UR (blue) and with UR (red). \textbf{(d)}: RE for the associated $T^{\text{PINN}}(S)$ for the cases without UR (blue) and with UR (red). In the legend, we show the value of the root mean square (rms) for each case.}
    \label{fig: TL holo 0.9}
\end{figure}

\begin{figure}[H]
    \centering
    \includegraphics[width = \textwidth]{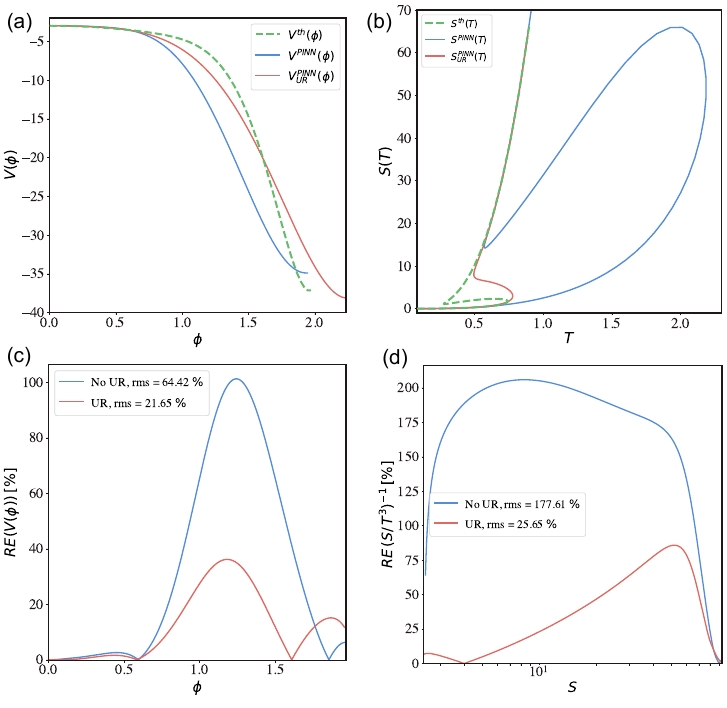}
    \caption{\textbf{Transfer learning of Einstein Field Equations for a highly-pronounced first order first transition}.$V(\phi)$ is the free form potential recovered by the NN (PINN) and the theoretical one (th). $S(T)$ is the thermodynamical curve (entropy as a function of temperature) recovered by the NN (PINN) and the theoretical one (th). RE refers to relative error between the NN model and the theoretical result. \textbf{(a)} Recovered functions $V^\text{PINN}(\phi)$ through TL without UR (solid blue), and with UR (solid red), compared with the theoretical function (dashed green), corresponding to $\phi_M=0.7$. \textbf{(b)} $S^{\text{PINN}}(T)$ relations associated to the recovered $V^\text{PINN}(\phi)$ without and with UR, compared to the theoretical one (dashed green). \textbf{(c)}: REs for the recovered $V^\text{PINN}(\phi)$ for the cases without UR (blue) and with UR (red). \textbf{(d)}: REs for the associated inverse function $(S^{\text{PINN}}/T^3)^{-1}$ (as a function of $S$) for the cases without UR (blue) and with UR (red). In the legend, we show the value of the root mean square (rms) for each case.}
    \label{fig: TL holo 0.7}
\end{figure}

As explained in the ``Multihead'' and the ``Training and Transfering'' subsections, we then use the frozen latent space to perform TL to stiffer regimes. In our approach, we do so for two new $S(T)$ curves corresponding to $\phi_M=0.9$ and $\phi_M=0.7$. We test UR on both cases and display our findings in figures \ref{fig: body holo eqs} (training of the \textit{body}), \ref{fig: TL holo 0.9} (case with $\phi_M=0.9$) and \ref{fig: TL holo 0.7} (case with $\phi_M=0.7$).

Observe that in this configuration, the UR possesses an extra dimension, making it three-dimensional. The first dimension is due to the independent variable $u$, the second dimension is associated with the $i$-index, which labels each BC along the path $S(T)\rightarrow\left\{T_i,S_i\right\}$, and the third dimension is linked with the \textit{stiffness parameter} $\phi_M$, which parametrizes the various morphologies of $S(T)$. The top panels of Fig. \ref{fig: body holo eqs} illustrate a slice of $\sqrt{|g|}$ perpendicular to the $\phi_M$ direction, representing a particular $S(T)_{\phi_M}$ or $V(\phi;\,\phi_M)$.

Before we present the results, we summarize the architecture and training procedure for the EFE case. These procedures are identical in scenarios both with and without UR to ensure a fair comparison of their performance. Each solution function in $\Vec{\psi}(u)=~\left( \tilde{\Sigma}(u),\tilde{A}(u),\phi(u),\nu_\Sigma(u),\nu_A(u),\nu_\phi(u) \right)$ is produced by an individual FCNN. The architecture of these NNs consists of $[32,32,32,32,128]$ layers for the main \textit{bodies} and $[16,16]$ for the five \textit{heads}, all utilizing \texttt{Tanh} activation. The free function $V^\text{PINN}(\phi)$ is derived from another FCNN with a \textit{body} structured as $[32,32,32,128]$ and five \textit{heads} composed of $[64,64]$, with \texttt{SiLU} as the activation for both the \textit{body} and \textit{heads}. The optimizer used for \textit{body} training and TL is \texttt{adam}, starting with an initial learning rate of $1\cdot 10^{-3}$. As part of the scheduling process, the learning rate is reduced by $1.5 \, \%$ every $5000$ epochs, both in training the body and in TL. In scenarios with UR, it is imposed in the loss calculation every 500 iterations, with a global coefficient $\lambda=5\cdot 10^{-8}$.

TL under UR exhibited significant improvement in recovering the free function $V(\phi;\phi_M)$ and the associated $S(T)_{\phi_M}$. Specifically, for TL to $\phi_M=0.9$ (Fig. \ref{fig: TL holo 0.9}), our UR approach accomplished a twofold improvement in the recovery of $V(\phi,0.9)$, reducing root mean square (rms) from roughly $5\%$ to about $3\%$. Consequently, this also signified a twofold improvement in $S(T)_{0.9}$, with rms decreasing from around $13\%$ to about $6\%$. In the context of $\phi_M=0.7$, results are less precise overall due to significant increases in problem stiffness. Nevertheless, there is a threefold enhancement through UR in retrieving $V(\phi,0.7)$, with rms dropping from approximately $64\%$ to approximately $22\%$, and a sevenfold improvement in $S(T)_{0.7}$, reducing rms from around $178\%$ to roughly $26\%$ (Fig. \ref{fig: TL holo 0.7}). In both scenarios, UR demonstrates improvements in recovery, and for $\phi_M=0.7$, it enables a closer TL approximation, whereas the conventional MH case resulted in extraordinarily large errors.


\section{Conclusions} \label{sec: conclusions}

We have presented two techniques to solve nonlinear, multiscale differential equations (DEs) with Physics-Informed Neural Networks (PINNs), placing special emphasis on inverse problems.
These techniques, which we term multi-head (MH) and Unimodular Regularization (UR), have proved effective in various examples of nonlinear, multiscale DEs. We have shown that, by using a single NN structure (the \textit{body}) shared across multiple NN substructures (\textit{heads}), MH training can learn the broader space of solutions for variations of the DE.
We verified this approach using variations in the DEs’ parametric dependence, initial conditions (ICs), and boundary conditions (BCs). The learned general solution space (or latent space) can then be leveraged to find specific solutions for new DE variations under more extreme or stiffer regimes, even if the MH \textit{body} was not initially trained for these cases.
This approach is known as transfer learning (TL). Furthermore, we have developed Unimodular Regularization (UR), a technique that imposes specific conditions on the determinant of the metric tensor induced by embedding the latent space into the input/parameter space.
This idea is grounded in differential geometry and draw inspiration from General Relativity. The imposed conditions quantify and regulate the latent space’s sensitivity to specific variations in the DEs. We have found that combining the MH approach with UR in the latent space of solutions significantly enhances PINN efficiency, facilitating the TL process and thus enabling more effective solution discovery.

Our results are derived by implementing the MH technique both with and without UR, then comparing the solutions obtained to a numerical baseline. We apply this approach to three distinct sets of ordinary differential equations (ODEs): the flame equation, the van der Pol (damped) oscillator, and the Einstein Field Equations (EFEs), being the last an inverse problem.

For the flame equation, a nonlinear first-order ODE, UR reduces the root mean square error (RMS) from 35.1\% to 0.5\%.
For the van der Pol oscillator, a nonlinear second-order ODE, UR enables TL to highly stiff regimes, resulting in a reduction of the RMS from 52.4\% to 5.7\%.
Finally, for the Einstein Field Equations (EFE)—an inverse, nonlinear system of first-order ODEs—we can simultaneously recover an approximate free-form function that appears in the DEs, together with the corresponding solutions.
The free function is inferred from its physically motivated correspondence to a physical relation between different BCs (an implicit mapping in the equations). In this final case, we applied TL to two increasingly stiff regimes. The first transfer showed a factor-of-2 improvement with UR. The second demonstrated a factor-of-3 improvement—or factor-of-7 when comparing the recovered free-form function or the corresponding BC relation. In this second instance, only approximate recovery of the full solution’s morphology is achieved when UR is included.

Although we have only considered ODEs, the extension of our setup to Partial Differential equations (PDEs) is almost straightforward. In the presented setup, promoting the system from ODEs to PDEs by adding a new independent variable is the same as considering additional Bundle parameters, since in both cases these are just inputs to the model. However, this convenience comes at a cost: for higher-dimensional PDEs or bundle solutions, we must sample the expanded input space on a multidimensional grid, which raises the effective batch size and correspondingly increases training time.

Furthermore, applying the UR method has some disadvantages and computational limitations. The first one is the training time. By including UR one needs to compute additional gradients during training in order to compute the metric tensor of the latent space. This increases the training time by $5\%-10 \%$, but this number is highly dependent on $N$ and the batch size. Second, storing the metric tensor and its derivatives for every batch element can exhaust GPU memory (in our final EFE experiment, adding one extra head exceeded the capacity of an NVIDIA A30 GPU). To address these issues, future work will explore low‐rank or sparse tensor approximations, gradient checkpointing or selective gradient retention, and optimized CUDA kernels to reduce both memory usage and runtime.

We have provided three examples demonstrating that, with these algorithms, it is possible to tackle very complex problems in various scientific fields where unknown phenomena must be discovered through the inversion of differential equations, as shown in the last provided example. In the future, we expect to apply this method to two different situations. The first one is to try to recover the potential in the holography setup in the most stiff cases. This will allow us to access physical situations with a very rich phenomenology. The second one is to apply the MH setup to gain some insights about the phenomena of turbulence. However, as we have explained, going directly to the Navier-Stokes equation is both computationally and conceptually demanding. It is for this reason that we are currently applying the set-up first to simpler PDEs, such as Burger's equation. This will allow us to recover the basis elements that the NN is choosing to solve the equation, up to a certain precision. More generally, we consider these algorithms could be extremely powerful and generalizable to a wide range of areas, including General Relativity, complex fluid dynamics, social sciences, and beyond.


\section{Author contribution}
All the authors have contributed equally to this work. Specifically all the authors have actively participated on the conceptual idea, writing code, training models and writing the final manuscript.

\section{Data Availability}
The data points used for BCs in subsection ``Einstein Field Equations'' are available in the GitHub repository \url{https://github.com/pedrota2000/Unimodular_regularization}. The trained models are available upon request to the authors.

\section{Code availability}
The \texttt{Python} codes used to generate the results are available in the GitHub repository \url{https://github.com/pedrota2000/Unimodular_regularization}.

\section{Competing Interests}
The authors declare no competing interests.

\section*{Acknowledgements}

PT is supported by the project “Dark Energy and the Origin of the Universe” (PRE2022-102220), funded by MCIN/AEI/10.13039/501100011033. Funding for the work of PT, PT and RJ was partially provided by
project PID2022-141125NB-I00, and the “Center of Excellence Maria de Maeztu 2020-2023” award to the
ICCUB (CEX2019- 000918-M) funded by MCIN/AEI/10.13039/501100011033.

\newpage

\section*{Supplementary Note 1: Mathematical form of the metric for UR and comparison to Jacobian Regularization}\label{ap: math of metric}\label{appA}

In this section, we mathematically show that a Jacobian regularization\footnote{Note that we changed notation in the choice of index in $J_{c,i}(\mathbf{x})$ from \cite{jacobianreg}, to $J_{i,\mu}(\mathbf{x})$ to make contact with the index notation in the subsection ``Unimodular Regularization'' within the ``Methods'' section, and throughout this work.} (JR) of the Frobenius norm of the pseudo-Jacobian \cite{jacobianreg}:
\begin{equation}
	L_\text{JR} = \lambda_\text{JR}\sum_\text{batch} ||J(x)||^2_F = \lambda_\text{JR}\sum_\text{batch} \sum_{i=1}^d\sum_{\mu=1}^n \left[J_{i,\mu}(\mathbf{x})\right]^2\,\,\,\,\,\,;\,\,\,\,\,J_{i,\mu}(\mathbf{x})\equiv\left.\frac{\partial H_i}{\partial x^\mu}\right|_{\mathbf{x}}
\end{equation}
is implicit in UR, with $n$ inputs and a dimension of the latent space $d$ (number of ``categories'' in \cite{jacobianreg}). This is because it appears as one of the terms in the computation of the determinant of the metric that is then regularized to unity, i.e.:
\begin{equation}
	L_\text{UR} = \lambda\sum_\text{batch} \left[\sqrt{g(x^\mu)} - \vec{\mathcal{I}}\right]^2\,.
\end{equation}

Before the computation, we make the following remark: the Frobenious norm of the pseudo-Jacobian matrix is defined such that all the partial derivatives with respect to inputs $x^\mu$ with $\mu=1,\dots,n$ are minimized individually. However, this scalar quantity lacks an interpretation in geometrical terms. In fact, we call the matrix $J_{i,\mu}\left(\textbf{x}\right)$ a pseudo-Jacobian because it is not generally a square matrix, and therefore cannot be the geometrical Jacobian as is usually understood in the literature. On the other hand, the minimization of the determinant of the metric,  $g=\text{det}(g_{\mu\nu})\rightarrow 1$ (through UR) 
has a clear geometrical interpretation; the determinant of the metric, or geometric Jacobian, measures the volume of a differential of hyper-volume in the latent space $H_i(x^\mu)$, where $i=1,\dots,d$, being $d$ the dimension of the latent space. Regularizing this quantity to unity leads to transforming the latent space such that this differential of hyper-volume is smoothly responsive to all inputs $x^\mu$, while keeping this response small if the change in $x^\mu$ is also small. This second point is achieved by minimizing the derivatives of the latent space with respect to all the inputs (similar to Jacobian regularization, but in a way that preserves the geometrical interpretation).\\

To show this, let us explicitly compute $g=\text{det}(g_{\mu\nu})$. The metric in our geometric construction is defined as:
\begin{equation}
	g_{\mu\nu} \,=\, \frac{\partial \vec{\Omega}}{\partial x^\mu}\cdot \frac{\partial \vec{\Omega}}{\partial x^\nu}
\end{equation}
being
$$\vec{\Omega} \,=\, \left(x^\mu, H_i(x^\mu)\right)=(x^1,\dots,x^n,H_1(x^\mu),\dots, H_d(x^\mu)) \vspace{0.1cm}$$
where $x^\mu\in \mathbb{R}^{n}$ are the inputs to the NN (independent variable, IC/BC, parameters), and $H\in\mathbb{R}^{d\times n}$ is the latent space given by the output of the last hidden layer of the body of the NN. This metric is computed for every point in the batch. \\

We denote the derivatives with respect to all inputs as:

\[
\frac{\partial}{\partial x^\mu} \equiv \left( \frac{\partial}{\partial x^1}, \ldots, \frac{\partial}{\partial x^n} \right) \equiv (\partial_1, \ldots, \partial_n)
\]

Then:
\[
\frac{\partial \bar{\Omega}}{\partial x^\mu} = 
\begin{pmatrix}
	\partial_1 \\
	\vdots \\
	\partial_n
\end{pmatrix}
\cdot
\left( x^1, \ldots, x^n, H_1(x^\mu), \ldots, H_d(x^\mu) \right)=
\begin{pmatrix}
	1 & \cdots & 0 & \partial_1 H_1  & \cdots & \partial_1 H_d \\
	
	\vdots & \ddots & \vdots & \vdots & \ddots & \vdots \\
	0 & \cdots & 1 & \partial_n H_1 & \cdots & \partial_n H_d
\end{pmatrix}_{n\times (n+d)}
\]

Computing then the metric in matrix form:
\begin{align}
	\left(g_{\mu\nu}\right)_{n\times n} = \left(\frac{\partial \vec{\Omega}}{\partial x^{\mu}}\right) \cdot \left(\frac{\partial \vec{\Omega}}{\partial x^{\nu}}\right)^T = 
	\begin{pmatrix}
		1 + \sum\limits_{i=1}^{d} a_{1i}^2 &  \cdots & \sum\limits_{i=1}^{d} a_{ni} a_{1i} \\
		\vdots & \ddots & \vdots \\
		\sum\limits_{i=1}^{d} a_{1i} a_{ni} & \cdots & 1 + \sum\limits_{i=1}^{d} a_{ni}^2 
	\end{pmatrix}
\end{align}
where we defined $a_{\mu i}\equiv \partial_\mu H_i$. This result can also be written as:
\begin{equation}
	g_{\mu\nu} = \left\{\begin{array}{lcl}
		1+\sum_{i=1}^d \left(\partial_\mu H_i\right)^2 &  ; & \mu=\nu\\ && \\
		\sum_{i=1}^d \left(\partial_\mu H_i\right)\left(\partial_\nu H_i\right) & ; & \mu\neq \nu
	\end{array}\right.
\end{equation}

The expression of the determinant for general $n,d$ becomes lengthy. For the purpose of this work, let us show an example for the case $n=2, \,d=2$, where we have the determinant:
\begin{align}
	g=\text{det}\left(g_{\mu\nu}^{(2,2)}\right) &= 1 + \left(\sum_{\mu=1}^n \sum_{i=1}^d a_{\mu i}^2 \right) + \left(a_{11}a_{22}-a_{12}a_{21}\right)^2 \notag \\
	& = 1 + \sum_{\mu=1}^n \sum_{i=1}^d\left[J_{i,\mu}(\mathbf{x})\right]^2 + \left(a_{11}a_{22}-a_{12}a_{21}\right)^2 = \notag \\ & \notag \\
	& \equiv 1 + J^2 + A^2 \label{eq: g for n=2, d=2}
\end{align}
where $A^2=\left(a_{11}a_{22}-a_{12}a_{21}\right)^2 > 0$. In the second line we used the definition of the matrix terms of the pseudo-Jacobian matrix used in \cite{jacobianreg}, $a_{\mu i}=J_{i,\mu}(\mathbf{x})\equiv \partial H_i/\partial x^\mu$. Thus, we can see that regularizing $g\rightarrow 1$ (we actually do $\sqrt{g}\rightarrow 1$) is equivalent to minimizing the factor $(J^2 + A^2)$, which is a regularization that includes JR (minimizing the Frobenius norm $J^2$), together with an extra regularizing factor.\\

This calculation can be generalized to arbitrary number of inputs $n$ and dimension of the latent space $d$.

\section*{Supplementary Note 2: Relation of UR with the Lipschitz constant.} \label{app: lipzsitz}
The Lipshitz constant $L$ of a function $f$ is defined as
\begin{equation*}
	\frac{|f(y)-f(x)|}{|y-x|}\leq L
\end{equation*}
When considering input to solution mapping $u(x^\mu)$ we can write the this condition as follows
\begin{equation*}
	\frac{|u(y^\mu)-u(x^\mu)|}{|y^\mu-x^\mu|} \leq L
\end{equation*}
It is easy to prove that this quantity is bounded by the 2-norm of the Jacobian of the function $u$
\begin{equation*}
	L\leq \sup_{x^\nu}\left|\left|\frac{\partial u(x^\nu)}{\partial x^\nu}\right|\right|
\end{equation*}
Note that by using the chain rule we can relate the derivative of $u$ with respect to the inputs to the derivative of the latent space as
\begin{equation*}
	\frac{\partial u(x^\mu)}{\partial x^\nu} \,=\,\sum_i \frac{\partial u(x^\mu)}{\partial H_i(x^\mu)}\cdot \frac{\partial H_i(x^\mu)}{\partial x^\mu}
\end{equation*}
since the determinant of $g$ includes the derivatives of the latent space with respect to the inputs (as shown in ``Supplementary Note 1'') making this quantity small would make the Lipschitz constant small.\\ 

Intuitively, we can think of this regularization techniques as follows. The determinant of the metric $g$ is proportional to the derivatives of the latent space with respect to the inputs. These derivatives are quantifying the sensitivity of the latent space to changes of independent variables and parameters of the differential equations. Let us consider two different limits: if the derivatives are very small, then the response of the latent space will be constant, and we will be learning no structure at all to project down to the actual solution. On the other hand, if the response of the latent space is very sensitive to the inputs, there will be too much structure in the latent space and it will be very difficult for the head to predict outside the range of training during the transfer learning. Thus, we will try to find the perfect balance between these two limits by playing with the hyperparameters $\lambda$ and $N$ (defined in the main text in the ``Unimodular Regularization'' subsection within the ``Methods'' section).


\end{document}